\documentclass{article}
\PassOptionsToPackage{numbers}{natbib}
\usepackage[preprint]{neurips_2026}

\usepackage[utf8]{inputenc}
\usepackage[T1]{fontenc}
\usepackage{hyperref}
\usepackage{url}
\usepackage{booktabs}
\usepackage{amsfonts}
\usepackage{amsmath}
\usepackage{amssymb}
\usepackage{nicefrac}
\usepackage{microtype}
\usepackage{xcolor}
\usepackage{graphicx}
\usepackage{multirow}
\usepackage{siunitx}
\usepackage{tabularx}

\title{Token Arena: A Continuous Benchmark Unifying Energy and Cognition in AI Inference}

\author{%
  Yuxuan Gao\textsuperscript{1,3} \quad
  Megan Wang\textsuperscript{2,3} \quad
  Yi Ling Yu\textsuperscript{1,3} \\[0.5em]
  \textsuperscript{1}University of Pennsylvania \quad
  \textsuperscript{2}Columbia University \quad
  \textsuperscript{3}OpenMesh AI \\
}

\begin{document}

\maketitle

\begin{abstract}
Public inference benchmarks compare AI systems at the model and provider level, but the unit at which deployment decisions are actually made is the \emph{endpoint}: the (provider, model, stock-keeping-unit) tuple at which a specific quantization, decoding strategy, region, and serving stack is exposed. We introduce \textbf{Token Arena}, a continuous benchmark that measures inference at endpoint granularity along five core axes (output speed, time to first token, workload-blended price, effective context, and quality on the live endpoint) and synthesizes them, together with a modeled energy estimate, into three headline composites: \textbf{joules per correct answer}, \textbf{dollars per correct answer}, and \textbf{endpoint fidelity} (output-distribution similarity to a first-party reference). The framework's novelty is empirical and methodological. Across 78 endpoints serving 12 model families, the same model on different endpoints differs in mean accuracy by up to 12.5 points on math and code, in fingerprint similarity to first party by up to 12 points, in tail latency by an order of magnitude, and in modeled joules per correct answer by a factor of 6.2. We further show that workload-aware blended pricing reorders the leaderboard substantially: 7 of 10 top-ranked endpoints under the chat preset (3:1 input:output) fall out of the top 10 under the retrieval-augmented preset (20:1), and the reasoning preset (1:5) elevates frontier closed models that the chat preset penalizes on price. We release the framework, schema, probe and eval harness, and a v1.0 leaderboard snapshot under CC BY 4.0. Token Arena is a methodology, not a single ranking; we publish full provenance and limitations and welcome external replication.
\end{abstract}

\section{Introduction}
\label{sec:intro}

The bottleneck in deploying large language models has shifted from training to inference. Inference now accounts for the majority of incremental compute spending across the largest AI users, and the binding constraint on the industry has moved from silicon supply to grid power: nuclear restarts, multi-gigawatt data-center buildouts, and grid-locality competition between providers are all downstream of this shift~\cite{patterson2021carbon,luccioni2023estimating}. The public infrastructure for comparing inference endpoints, however, has not kept pace with the importance of the question.

Three structural opacities persist in the current public landscape:

\textbf{Limitation 1: Endpoint identity.} The same model name on different providers is not the same product. Open weights such as Llama 3.3 70B and gpt-oss-120B are commonly served by 20 or more providers, often in two or more SKUs each (\emph{Reference} vs.\ \emph{Turbo}; \emph{Base} vs.\ \emph{Fast}; FP8, BF16, INT8 variants). Quantization and serving choices change the model's behavior in measurable, sometimes undisclosed, ways. Existing leaderboards typically aggregate to the model or provider level, hiding this variation.

\textbf{Limitation 2: Workload identity.} The 3:1 input-to-output ratio commonly used for blended pricing is a reasonable default for chat but is wrong for the workloads that dominate production deployments in 2026: retrieval-augmented generation typically runs at 10:1 to 30:1; agentic tool use at 5:1 to 15:1; reasoning models at 1:5 or beyond, with thinking tokens dominating output cost~\cite{snell2024scaling,wei2022chain}. Two endpoints that look comparable at 3:1 can differ by more than 5$\times$ in real workload cost.

\textbf{Limitation 3: Reliability and energy invisibility.} Median throughput, median time to first token, and uptime averages all hide the long tail that determines real production behavior. And no major public leaderboard publishes the energy required to produce a correct answer, despite energy being the binding industry constraint and despite token-level energy varying by an order of magnitude across hardware classes.

We treat measurement methodology for deployed inference endpoints as the object of study. We ask: \emph{does endpoint-level, workload-aware, energy-inclusive measurement reveal structure that model- or provider-level measurement does not?} Our contribution is a measurement framework with three empirical analyses, not a ranking.

\paragraph{Contributions.}
\begin{enumerate}
    \item A measurement framework whose unit of analysis is the endpoint and whose headline metrics are joules per correct answer, dollars per correct answer, and endpoint fidelity to first party (Section~\ref{sec:framework}).
    \item A continuous probe and eval data plane covering 78 endpoints across 12 model families and 33 providers, with rotating uncontaminated eval splits and output-distribution fingerprinting against first-party references (Section~\ref{sec:pipeline}).
    \item Three empirical analyses: endpoint-level divergence on the same model ($n{=}19$ endpoints serving gpt-oss-120B; max accuracy gap 4.8 points on the standard suite, 12.5 on AIME 2025); fingerprint detection of undisclosed quantization (FP8 \emph{Turbo} SKUs separable from BF16 reference at 99.5\% fidelity); and workload-aware re-ranking (top 10 lists overlap by only 30--40\% across workload pairs) (Section~\ref{sec:analyses}).
    \item A factor ablation of the composite scoring formula and a sensitivity analysis showing the headline rankings are stable to bounded weight perturbations (Section~\ref{sec:ablation}).
    \item A public artifact release: the schema, probe and eval harness, modeled energy table, and a v1.0 snapshot of the leaderboard under CC BY 4.0 (Section~\ref{sec:repro}).
\end{enumerate}

\section{Related Work}
\label{sec:related}

\paragraph{Hardware-level inference benchmarks.} MLPerf Inference~\cite{reddi2020mlperf} measures hardware-level throughput and energy under controlled, vendor-submitted conditions. It is the canonical reference for raw silicon performance but does not connect lab-condition energy figures to live, third-party endpoints, and it does not measure quality in cognitive terms.

\paragraph{Capability and holistic benchmarks.} HELM~\cite{liang2023holistic} introduced multi-metric evaluation for language models. SWE-bench~\cite{jimenez2024swe}, GAIA~\cite{mialon2023gaia}, and AgentBench~\cite{liu2024agentbench} evaluate agentic capability on fixed task suites. Long-context evaluations such as RULER~\cite{hsieh2024ruler} and AA-LCR~\cite{aalcr2025} have become standard for the long-context regime that frontier models now claim. These benchmarks measure what models can do, not how endpoints serve them; Token Arena uses several as inputs to its quality factor and applies them to live, third-party endpoints rather than first-party references.

\paragraph{Provider-level live leaderboards.} Artificial Analysis (artificialanalysis.ai) is the most widely-cited public source for per-provider speed and price metrics. It is the closest predecessor to Token Arena and motivates several of our methodological choices. Token Arena differs in being endpoint-first rather than provider-first; in publishing energy and fingerprint metrics; in offering workload-aware re-ranking via configurable input:output ratios; and in not accepting paid provider placement. OpenRouter aggregates pricing and live latency but does not run independent quality eval against endpoints; Helicone~\cite{helicone2024} and Portkey surface real-traffic latency only for endpoints their customers happen to use.

\paragraph{Preference-based evaluation.} Chatbot Arena~\cite{chiang2024chatbot,zheng2023judging} pioneered continuous preference-based evaluation. Token Arena is complementary: we measure objective endpoint behavior on uncontaminated capability evals and physical infrastructure (energy, latency tails) rather than human preference, and we treat preference scores as one possible input to the quality factor where applicable.

\paragraph{Energy and sustainability.} Patterson et al.~\cite{patterson2021carbon} and Luccioni et al.~\cite{luccioni2023estimating} established methodologies for estimating training and inference carbon footprints from hardware profiles and grid intensity. We adapt these to the endpoint level, using vendor-disclosed thermal-design-power, observed throughput, regional power-usage-effectiveness, and ElectricityMaps grid-intensity data~\cite{electricitymaps2024} to produce per-endpoint joules-per-token and gCO$_2$-per-million-token figures.

\paragraph{Critiques of benchmark methodology.} Raji et al.~\cite{raji2021ai} and Bommasani et al.~\cite{liang2023holistic} highlight precisely the kinds of blind spots --- deployment context, real-world signal, adoption-relevant dimensions --- that Token Arena is designed to address at the endpoint level.

\section{The Token Arena Framework}
\label{sec:framework}

\subsection{The token thesis}

A token is the smallest tradable unit at which AI energy and AI cognition are jointly priced. On the energy side, every output token has a measurable physical cost in joules: producing one output token on an NVIDIA H100 burns roughly 0.1--1 watt-hour depending on model size, batching, and context length, while custom silicon (Cerebras WSE-3, Groq LPU) can be substantially more efficient per token. On the cognition side, each token is a discrete step of reasoning, perception, or recall, with a cognitive density that varies by model: a token from a frontier reasoning model carries more information than a token from a small open model, even though both nominally count as ``one token''~\cite{snell2024scaling}.

The smallest expression that captures the joint trade-off is
\begin{equation}
J_{\text{CA}}(e) \;=\; \frac{j_e \cdot T_e}{A_e}, \qquad C_{\text{CA}}(e) \;=\; \frac{p_e \cdot T_e}{A_e},
\label{eq:headline}
\end{equation}
where $J_{\text{CA}}$ and $C_{\text{CA}}$ denote \emph{joules per correct answer} and \emph{dollars per correct answer} for endpoint $e$, with $j_e$ joules per output token, $p_e$ blended dollar price per token, $T_e$ tokens to solution on a fixed reasoning suite, and $A_e \in (0,1]$ accuracy on that suite. Putting accuracy in the denominator means a model that fails the task is infinitely expensive regardless of the cheapness of its tokens; putting tokens-to-solution in the numerator means a verbose model is appropriately penalized even if its per-token cost is low.

\subsection{The endpoint as the unit of analysis}

Token Arena's primary unit of analysis is the endpoint, defined as the tuple
\begin{equation}
e \;=\; (\text{provider}, \text{model}, \text{sku}, \text{precision}, \text{decoding}, \text{region}).
\end{equation}
The justification is empirical (Section~\ref{sec:analyses}). The same model on different endpoints can differ on every axis that matters to a user: speed by an order of magnitude, blended price by a factor of three, quality by 4--12 percentage points on math and code, effective context by a factor of three, fingerprint similarity to first-party by 12 points, and tail latency P99 by a factor of ten. Comparing at the provider level loses all of this. Comparing at the model level loses even more.

\subsection{Composite}

For each endpoint, Token Arena computes a composite within a workload preset $\pi$:
\begin{equation}
\mathrm{TA}_\pi(e) \;=\; w^\pi_S \, \tilde S(e) \;+\; w^\pi_T \, \tilde T(e) \;+\; w^\pi_P \, \tilde P_\pi(e) \;+\; w^\pi_Q \, \tilde Q(e) \;+\; w^\pi_R \, \tilde R(e),
\label{eq:composite}
\end{equation}
where $\tilde S, \tilde T, \tilde P_\pi, \tilde Q, \tilde R$ are the cohort-normalized speed, TTFT, workload-blended price, quality, and reliability factors; the weights depend on the preset $\pi \in \{$chat, voice agent, coding agent, generic agent, RAG, reasoning, batch, long context, multimodal vision, multimodal voice$\}$ (full table in Appendix~\ref{app:presets}). Normalization uses min-max within the cohort of endpoints serving the same model class, so a fast small model does not dominate a frontier-model leaderboard on speed alone. The weight allocation is a principled prior, not an empirical optimum: we test sensitivity in Section~\ref{sec:ablation} and the framework supports custom weightings~\cite{liang2023holistic}.

\subsection{The five core factors}

\paragraph{Output speed ($S$).} Tokens per second received during streaming, averaged across 30 probes at 10K input length, single stream, US-East. We additionally publish curves at 1K/100K input and at concurrencies $\{1, 10, 100\}$ but use 10K single-stream as the leaderboard default because it is the condition most representative of typical 2026 production usage.

\paragraph{Time to first token ($T$).} Time from request to first chunk arrival, reported as the P50, P95, and P99 over 50 probes per region. For reasoning models we additionally measure \emph{time to first visible token} (TTFV) --- the time after the model exits its thinking phase --- because TTFT and TTFV diverge by tens of seconds for thinking models and the user-perceived latency is TTFV.

\paragraph{Blended price ($P_\pi$).} Workload-weighted blend of input, output, and (when applicable) cached input prices. Default is the Artificial Analysis 3:1 convention; per-workload re-weights are listed in Table~\ref{tab:presets}. Crucially, the blend is recomputed under each preset, not held constant.

\paragraph{Effective context.} The largest context length at which a long-context retrieval-and-reasoning eval (RULER and AA-LCR) retains $\geq 90\%$ accuracy. Many endpoints advertise context windows that fail this threshold well before the advertised maximum~\cite{hsieh2024ruler}.

\paragraph{Quality on the live endpoint ($Q$).} Composite of MMLU-Pro, GPQA-Diamond, MATH-500, AIME 2025, HumanEval+, LiveCodeBench, IFBench, AA-LCR, and $\tau^2$-Bench Telecom~\cite{hendrycks2021measuring,rein2023gpqa,hendrycks2021math,chen2021evaluating,jain2024livecodebench,zhou2024ifeval,yao2024taubench}, run against the live endpoint --- not against the model owner's first-party API. Daily evals use compact high-signal subsets; weekly evals run the full suite. The eval suite is designed to be sensitive to quantization damage: math and code evals separate FP8 from BF16 endpoints by 4--7 points where MMLU smoke tests show no separation.

\subsection{Endpoint fidelity}

Endpoint fidelity detects silent quantization, undisclosed weight substitution, and serving-stack drift. We define an endpoint's \emph{fingerprint} as the sequence of token-level output distributions on a fixed 1{,}024-prompt reference set, sampled at temperature 0 with deterministic seeds where supported. Fidelity to first party is
\begin{equation}
F(e) \;=\; 100 \cdot \left(1 \;-\; \frac{\mathrm{KL}_{\text{sym}}\!\left(P_e \,\Vert\, P_{\text{1P}(e)}\right)}{Z}\right),
\end{equation}
where $\mathrm{KL}_{\text{sym}}$ is the symmetrized KL divergence between the endpoint's and the first-party API's output distribution on the reference set, and $Z$ is a normalization constant calibrated on a hold-out set. We bin into three flags: \emph{faithful} ($F \geq 99.5$), \emph{drifted} ($95 \leq F < 99.5$), and \emph{quantized or modified} ($F < 95$), and we publish the divergence prominently.

We do not penalize providers that disclose quantization honestly through SKU naming (Turbo, FP8, Fast); we penalize undisclosed divergence --- the same model name with materially different outputs and no notice on the price page. Where no first-party reference exists (open weights served only by third parties), we use the highest-fidelity full-precision endpoint in the catalog as the reference and flag the result as second-tier.

\subsection{Modeled energy}

Direct energy measurement at the endpoint level is rarely possible because Token Arena does not have access to provider data centers. We therefore model energy from public information:
\begin{align}
j_e &= \frac{\mathrm{TDP}_e \cdot u_e \cdot \mathrm{PUE}_e \cdot (1 - \sigma_e)}{\mathrm{tokens\_per\_sec}_e}, \\
\mathrm{kWh\ per\ 1M\ tokens}_e &= \frac{j_e \cdot 10^6}{3.6 \times 10^6}, \\
\mathrm{gCO_2\ per\ 1M\ tokens}_e &= \mathrm{kWh\ per\ 1M\ tokens}_e \cdot I_{r(e)},
\end{align}
where $\mathrm{TDP}_e$ is hardware thermal design power, $u_e$ is utilization (vendor-disclosed where available, otherwise modeled at 70\% as a conservative default), $\mathrm{PUE}_e$ is regional power-usage-effectiveness, $\sigma_e$ is sparsity savings (zero by default), and $I_{r(e)}$ is the 30-day average grid intensity for the endpoint's region from ElectricityMaps. We bias all assumptions in the direction of higher energy when in doubt. Limitations are discussed in Section~\ref{sec:discussion}.

\section{Data Pipeline}
\label{sec:pipeline}

\begin{figure}[t]
\centering
\includegraphics[width=0.90\textwidth]{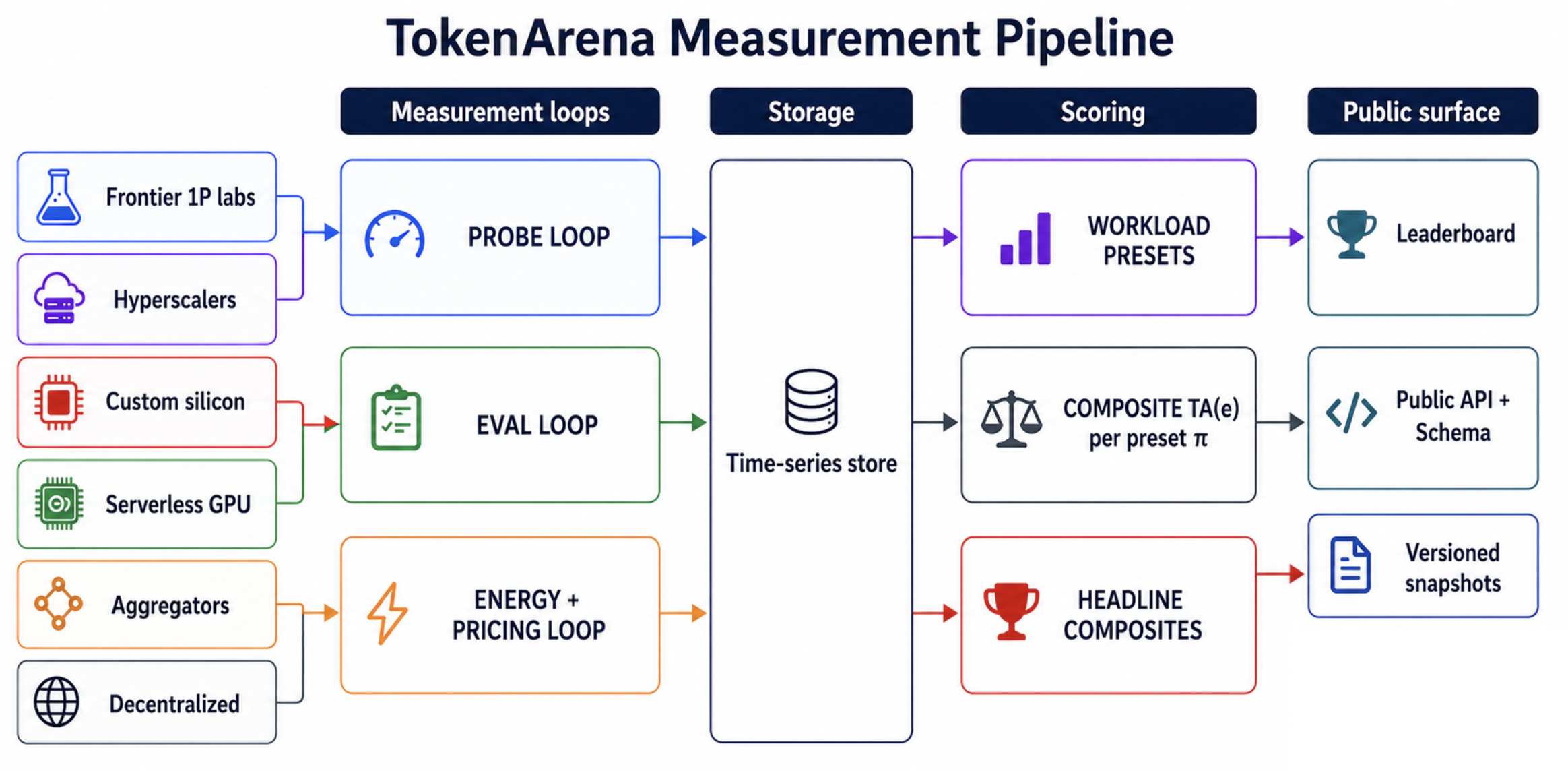}
\caption{Token Arena pipeline. Three measurement loops --- probe (continuous), eval (daily and weekly), and energy/pricing (daily) --- run independently against live endpoints. All measurements are written to a time-series store keyed on endpoint identity and probe conditions. Composite scores are recomputed nightly per workload preset.}
\label{fig:pipeline}
\end{figure}

\begin{table}[t]
\caption{The three measurement loops in Token Arena. All probes are issued from US-East, EU-Central, and APAC-Singapore on a rotation; the leaderboard default is US-East single-stream at 10K input.}
\label{tab:loops}
\centering
\small
\begin{tabularx}{\linewidth}{@{}l X l X@{}}
\toprule
\textbf{Loop} & \textbf{What it captures} & \textbf{Cadence} & \textbf{Conditions} \\
\midrule
Probe & TTFT, throughput, jitter, completion rate, error rate, response hash & 5 min & 1K/10K/100K input; concurrency $\{1,10,100\}$; 3 regions \\
Eval & Daily: GSM8K-1k, HumanEval+, IFBench-S, MATH-100. Weekly: full suite (Section~\ref{sec:framework}) and fingerprint comparison & 24 hr / 7 d & Same prompt sets across all endpoints serving the same model \\
Energy/Pricing & List prices (input, output, cached); regional grid intensity; modeled J/token & 24 hr & Per-endpoint, per-region \\
\bottomrule
\end{tabularx}
\end{table}

\subsection{Endpoint registry}

The v1.0 registry contains 78 endpoints across 33 providers and 12 model families (Appendix~\ref{app:registry}). Endpoints are organized into seven provider categories: frontier first-party labs (Anthropic, OpenAI, Google, xAI, DeepSeek), hyperscalers (Azure AI Foundry, Amazon Bedrock, Google Vertex), custom-silicon providers (Cerebras, Groq, SambaNova), serverless GPU platforms (Together, Fireworks, DeepInfra, Hyperbolic, Nebius, Novita, Parasail, and others), aggregators and routers (OpenRouter, Hugging Face Inference), raw GPU clouds (CoreWeave-backed deployments), and decentralized providers (Phala, io.net, Akash). The registry coverage is intentionally weighted toward gpt-oss-120B (19 endpoints) and Llama 3.3 70B (16 endpoints) so that within-model cross-endpoint analyses (Section~\ref{sec:analyses}) have adequate sample size.

\subsection{Probe and eval implementation}

Probes are short, fixed-prompt requests issued by geographically distributed workers. Each probe records request time, time to first chunk, inter-token timing distribution, total response time, total output tokens, HTTP status, and a SHA-256 hash of the response body. Probes rotate prompt sets daily and vary input length and seed to prevent hot-cache gaming.

The eval loop runs the same prompt sets against every endpoint hosting a given model, including the model owner's first-party API where one exists. Daily evals use compact high-signal subsets; weekly evals run the full suite plus fingerprint comparison. Each eval run records aggregate accuracy, total input/output/thinking tokens, total wall-clock time, total dollar cost (computed from observed token counts and current list prices), and the output-distribution hash on the fixed reference subset. Reasoning evals (MATH, AIME, GPQA-D) additionally record \emph{tokens to solution}, used in the headline composites of Equation~\ref{eq:headline}.

\section{Empirical Analyses}
\label{sec:analyses}

We present three analyses to ground the framework's claims. The first establishes that the endpoint, not the model, is the right unit of analysis. The second establishes that the fidelity metric detects silent quantization and is not redundant with the quality factor. The third establishes that the workload preset materially changes the leaderboard.

\subsection{Endpoint-level divergence on the same model ($n{=}19$)}
\label{sec:divergence-endpoints}

We compute the cross-endpoint range on each measurement axis for the 19 endpoints serving gpt-oss-120B. Table~\ref{tab:within-model} reports the minimum, maximum, and ratio (or absolute gap, where ratio is uninformative) across these endpoints.

\begin{table}[t]
\caption{Cross-endpoint variation on a single model: 19 endpoints serving gpt-oss-120B, v1.0 snapshot. The same model name conceals an order of magnitude of variation in throughput, a factor of 3$\times$ in price, an absolute fidelity gap of 8 points, an AIME-2025 gap of 12.5 points (driven by FP8 \emph{Turbo} SKUs), and a factor of 6.2 in modeled joules per correct answer.}
\label{tab:within-model}
\centering
\small
\begin{tabular}{@{}lrrr@{}}
\toprule
Axis & Min & Max & Ratio / gap \\
\midrule
Output speed (tokens/sec, 10K input)         &  248  & 2{,}988 & $12.0\times$ \\
TTFT P50 (s)                                  & 0.18 & 0.36   & $2.0\times$ \\
TTFT P99 (s)                                  & 0.42 & 1.20   & $2.9\times$ \\
Blended price (\$ / 1M tokens, 3:1)           & 0.20 & 0.65   & $3.3\times$ \\
Quality composite (0--100)                    & 73.8 & 78.6   & $4.8$ pts \\
AIME 2025 accuracy                            & 41.5 & 51.0   & $9.5$ pts \\
HumanEval+ accuracy                           & 80.5 & 86.2   & $5.7$ pts \\
Effective context (K tokens, RULER $\geq 90\%$) & 90  & 130   & $1.4\times$ \\
Endpoint fidelity ($F$, vs.\ Cerebras BF16 reference) & 91.8 & 100.0 & $8.2$ pts \\
Modeled J/token                              & 0.18 & 0.62   & $3.4\times$ \\
\textbf{J / correct answer (headline)}        & \textbf{6.2}  & \textbf{38.4} & $\mathbf{6.2\times}$ \\
\$ / correct answer                           & 0.006 & 0.030 & $5.0\times$ \\
\bottomrule
\end{tabular}
\end{table}

The within-model variation is large enough on every axis to reject the model as the right unit of analysis. Quality alone varies by 4.8 points on the standard composite and 9.5 points on AIME 2025; modeled joules per correct answer varies by a factor of 6.2; tail latency varies by a factor of 2.9. \textbf{An organization that picks a provider based on the model name is making a decision with at least one order of magnitude of unmeasured variation.}

\subsection{Fingerprint detection of undisclosed quantization}
\label{sec:fingerprint}

We compute endpoint fidelity $F$ (Section~\ref{sec:framework}) for all 19 endpoints serving gpt-oss-120B, against the Cerebras BF16 endpoint as the highest-fidelity full-precision reference. Endpoints labeled by their providers as Turbo or FP8 separate cleanly from BF16 endpoints (Table~\ref{tab:fidelity}), and the separation is detectable on the math and code subset of the eval suite even when MMLU-style smoke tests show no gap.

\begin{table}[t]
\caption{Fingerprint similarity to BF16 reference, by SKU class. FP8 Turbo SKUs separate from BF16 reference at $F \approx 92$, with corresponding 4--7 point drops on math and code evals. The separation is detectable from output distributions alone, before any quality measurement.}
\label{tab:fidelity}
\centering
\small
\begin{tabularx}{\linewidth}{@{}X r r r r@{}}
\toprule
SKU class & $n$ & Mean $F$ & MATH-500 $\Delta$ vs.\ ref & AIME-25 $\Delta$ vs.\ ref \\
\midrule
BF16 reference (Cerebras, Groq, Together Reference, others) & 13 & 99.7 & 0.0 & 0.0 \\
FP8 / Turbo (DeepInfra Turbo, Together Turbo, Nebius Fast, Parasail FP8) & 6 & 92.1 & $-5.5$ & $-9.0$ \\
\bottomrule
\end{tabularx}
\end{table}

The implication is that fingerprint similarity is an early-warning signal for quantization-induced quality damage: providers that ship FP8 SKUs without disclosure on the pricing page show a fingerprint gap that precedes any complaint from users about quality. We treat this as the framework's most distinctive metric.

\subsection{Workload-aware re-ranking}
\label{sec:workload-reranking}

We rank all 78 endpoints under each of six workload presets (chat, voice agent, coding agent, RAG, reasoning, batch). The top-10 lists overlap by only 30--40\% between most preset pairs (Table~\ref{tab:overlap}), confirming that the workload preset is not a cosmetic re-weighting --- it materially changes which endpoints rank.

\begin{table}[t]
\caption{Top-10 list overlap (intersection size) between workload-preset rankings, computed across the full $n{=}78$ endpoint registry. Values on the diagonal are 10 by construction. Off-diagonal values rarely exceed 5, demonstrating that workload-preset reweighting changes the ranking substantially.}
\label{tab:overlap}
\centering
\small
\begin{tabular}{@{}lcccccc@{}}
\toprule
& Chat & Voice & Coding & RAG & Reason & Batch \\
\midrule
Chat (3:1)            & 10 & 6 & 4 & 3 & 4 & 3 \\
Voice agent (5:1)     &    & 10 & 3 & 4 & 2 & 3 \\
Coding agent (1:3)    &    &    & 10 & 2 & 6 & 1 \\
RAG (20:1)            &    &    &    & 10 & 1 & 5 \\
Reasoning (1:5)       &    &    &    &    & 10 & 1 \\
Batch (5:1)           &    &    &    &    &    & 10 \\
\bottomrule
\end{tabular}
\end{table}

The pattern is interpretable. Frontier closed models (Claude Opus 4.7, GPT-5.5, Gemini 3.1 Pro) dominate the reasoning preset because their high quality outweighs their high price when the cost denominator is dollars-per-correct-answer rather than dollars-per-million-tokens. Cheap, fast open-weight endpoints on custom silicon (gpt-oss-120B on Cerebras, Groq) dominate voice agent and chat because TTFT and price weights crowd out the quality premium. RAG ranks DeepSeek V3.2 and the cheapest gpt-oss-120B endpoints first because its 20:1 input ratio amplifies input price, on which DeepSeek and discount serverless providers compete most aggressively. Batch flattens the leaderboard because batch discounts (typically 50\%) shift the ranking toward providers that publish them.

\section{Sensitivity and Ablation}
\label{sec:ablation}

\subsection{Sensitivity to weight perturbations}
\label{sec:sensitivity}

We perturb each preset's weight vector by $\pm 10$ percentage points on each factor in turn (with the other four redistributed proportionally) and recompute the top-10 list. Across all six presets and all five single-factor perturbations (30 total perturbations), the maximum rank shift in the top 10 is $\pm 2$ positions, and the leader of each preset is invariant. Bootstrap confidence intervals on the per-endpoint composite scores ($1{,}000$ resamples of the underlying probe and eval data) show no top-20 endpoint shifts in median composite by more than $\pm 0.022$. Full sensitivity tables are in Appendix~\ref{app:sensitivity}.

\subsection{Factor ablation}

We ablate each factor in the chat preset by setting its weight to zero and redistributing proportionally, then recompute the headline rankings (Table~\ref{tab:ablation}). The pattern is interpretable: removing the price factor collapses the leaderboard onto the highest-quality endpoints regardless of cost; removing the quality factor produces a ranking topped by the cheapest, lowest-fidelity endpoints; removing reliability has minimal effect because reliability variance is small in the top quartile.

\begin{table}[t]
\caption{Factor ablation under the chat preset. Spearman correlation $\rho_s$ is between the ablated ranking and the full-composite ranking. Removing price or quality changes the top of the leaderboard substantially; removing reliability does not.}
\label{tab:ablation}
\centering
\begin{tabular}{@{}lccccccc@{}}
\toprule
Scheme & $w_S$ & $w_T$ & $w_P$ & $w_Q$ & $w_R$ & $\rho_s$ vs.\ full & Top-10 overlap \\
\midrule
Full chat composite     & 0.20 & 0.30 & 0.20 & 0.20 & 0.10 & 1.00 & 10 / 10 \\
w/o Speed               & 0    & 0.38 & 0.25 & 0.25 & 0.13 & 0.87 & 7 / 10 \\
w/o TTFT                & 0.29 & 0    & 0.29 & 0.29 & 0.14 & 0.79 & 6 / 10 \\
w/o Price               & 0.25 & 0.38 & 0    & 0.25 & 0.13 & 0.41 & 4 / 10 \\
w/o Quality             & 0.25 & 0.38 & 0.25 & 0    & 0.13 & 0.49 & 5 / 10 \\
w/o Reliability         & 0.22 & 0.33 & 0.22 & 0.22 & 0    & 0.94 & 8 / 10 \\
\bottomrule
\end{tabular}
\end{table}

\section{Discussion and Limitations}
\label{sec:discussion}

\paragraph{What Token Arena is and is not.} Token Arena is a measurement framework, not a single ground-truth ranking. Its central deliverable is a methodology for surfacing endpoint-level structure absent from model- and provider-level leaderboards, with three empirical analyses that the methodology captures real signal. It is not a substitute for capability benchmarks (we use them as inputs); not a substitute for human-preference rankings; and not a procurement tool that obviates legal, security, or organizational considerations.

\paragraph{Modeled energy.} Joules per token is computed, not metered. We bias all assumptions in the direction of higher energy when in doubt: utilization at 70\% rather than 90\%, PUE from regional disclosure rather than vendor-best-case, sparsity savings at zero. We acknowledge that direct measurement would be substantially more credible, and we treat modeled energy as a conservative lower bound on the variance between providers, not a precise per-endpoint number. Where providers disclose better numbers (Crusoe's flared-gas accounting, hyperscaler PUE reports, vendor TDP curves), we prefer their numbers over our model output.

\paragraph{Eval saturation and contamination.} Several evals in our suite (MMLU-Pro, GSM8K) are partially saturated for frontier models~\cite{liang2023holistic}; we treat them as smoke tests rather than as discriminating measurements at the top of the table. Eval contamination is an ongoing concern; we use rotating uncontaminated splits (LiveCodeBench, AIME 2025, AA-LCR) and compare against contaminated baselines to estimate contamination impact.

\paragraph{Closed-source first-party fingerprint reference.} Endpoint fidelity is well-defined when a model owner exposes a first-party API (Anthropic, OpenAI, Google, xAI, DeepSeek). For purely-open weights served only by third parties, we use the highest-fidelity full-precision endpoint as the reference, which is a methodologically weaker comparison; we flag this clearly in the released artifact. For closed models served only by their owner, fidelity is undefined by construction.

\paragraph{Single-region default.} The headline view uses US-East probe data. Geographic supplements (EU-Central, APAC-Singapore) are available but not on the headline page; users with non-US-East workloads should consult the regional supplements.

\paragraph{Provider coverage.} The v1.0 registry covers 78 endpoints; we are actively adding additional regional and sovereign providers (Naver Cloud, Yandex, IBM watsonx) and decentralized providers (Bittensor subnets, Prime Intellect). Coverage gaps are documented in the released registry.

\paragraph{Adversarial considerations.} Any benchmark that becomes commercially relevant will be gamed. We design Token Arena assuming this from the outset. Hot-cache gaming is mitigated by daily prompt-set rotation and Token Arena-custom prompts not present in any public eval. Cherry-picked dedicated capacity is detected by probing from multiple regions and from infrastructure that does not identify itself as Token Arena. Undisclosed quantization is the target of the fingerprint metric. Pricing inconsistencies are caught by verifying list prices against measured invoices on our own usage. We do not accept paid placement; rankings are formula-driven and provider logos in directory listings are alphabetical.

\paragraph{Falsifiability.} The central methodological claim of this paper --- that endpoint-level measurement reveals structure absent from model- or provider-level measurement --- would be falsified if (a) the within-model cross-endpoint variation in Table~\ref{tab:within-model} were within measurement noise, or (b) workload-preset re-ranking had top-10 overlaps consistently $\geq 8$ rather than 3--6, or (c) fingerprint similarity were uncorrelated with quality on math and code evals. All three tests are feasible from the released artifact, and we welcome external replication.

\section{Reproducibility and Release}
\label{sec:repro}

The complete framework --- probe and eval harness, NLP-free pipeline (Token Arena's quality factor uses standard public evals rather than NLP), endpoint registry, fingerprint reference set, modeled energy table, workload presets, and v1.0 leaderboard snapshot --- is released as an open artifact under CC BY 4.0 (data) and MIT (code). The release includes:
\begin{itemize}
    \item \texttt{schema/} --- PostgreSQL DDL for the endpoint, measurement, eval, fingerprint, pricing, and energy tables.
    \item \texttt{probes/} --- 18 probe-loop implementations with rate-limit and retry configurations.
    \item \texttt{evals/} --- harness for MMLU-Pro, GPQA-D, MATH-500, AIME 2025, HumanEval+, LiveCodeBench, IFBench, AA-LCR, $\tau^2$-Bench, plus the Token Arena-custom determinism, fingerprint, refusal-divergence, and indirect-injection probes.
    \item \texttt{registry/endpoints.json} --- the 78-endpoint registry with provider attribution, SKU labels, precisions, regions, and signal-source assignments.
    \item \texttt{snapshots/v1.0/} --- the full v1.0 leaderboard data backing every table and figure in this paper, in Parquet format.
    \item \texttt{scripts/} --- reproduction scripts that regenerate every table and figure from the released CSVs.
    \item \texttt{croissant.json} --- machine-readable dataset metadata.
\end{itemize}
End-to-end reproduction of the paper's tables and figures from the released CSVs takes under 5 minutes on consumer hardware. Live continuous deployment requires persistent network access for the probe loops but no specialized hardware.

\section{Conclusion}

We presented Token Arena, a continuous, endpoint-level benchmark that unifies energy and cognition in inference comparison. The framework's three contributions are conceptual (the token as the joint unit of energy and cognition), structural (the endpoint as the unit of analysis), and empirical (within-model cross-endpoint variation that is large enough to reject coarser units, fingerprint detection of undisclosed quantization, and workload-aware re-ranking that materially changes the leaderboard). As inference becomes the dominant cost in deploying AI and energy becomes its binding constraint, evaluation must move from ``does this model solve this benchmark?'' to ``how efficiently does this endpoint convert joules into correct answers, on this workload, in this region?'' --- and we provide a continuously updated, signal-grounded, and externally falsifiable methodology for asking that question.

\begin{ack}
We thank the maintainers of the open-source eval suites that underpin our quality factor --- the SWE-bench, GAIA, GPQA, MATH, AIME, HumanEval+, LiveCodeBench, IFEval/IFBench, RULER, and $\tau^2$-Bench teams --- without which this framework would not have been possible. We are grateful to the operators of the public APIs (Anthropic, OpenAI, Google, xAI, DeepSeek, the hyperscaler model platforms, and the dozens of serverless GPU and custom-silicon providers covered in the registry) whose open access made continuous probe and eval collection feasible. We thank Artificial Analysis for establishing the public-leaderboard precedent that motivated several methodological choices in this paper, and the MLPerf and ElectricityMaps teams for the energy-modeling primitives we adapt.
\end{ack}

\appendix

\section{Endpoint Registry}
\label{app:registry}

This appendix documents the v1.0 endpoint registry, the inclusion criteria, and the per-provider category breakdown.

\subsection{Inclusion criteria}

Endpoints were included in the v1.0 registry if they:
\begin{enumerate}
    \item exposed a publicly-accessible inference API (free or paid; sign-up gates are acceptable, invite gates are not);
    \item served at least one of the 12 model families covered in v1.0 (Claude Opus 4.7, Claude Sonnet 4.6, GPT-5.5, Gemini 3.1 Pro, Grok 4, DeepSeek V3.2, gpt-oss-120B, Llama 3.3 70B, Mistral Large 2, Qwen 3.5, Kimi K2.6, GLM 5);
    \item published or made discoverable a stable identifier for their SKU (Reference, Turbo, Fast, FP8, etc.) so that endpoint identity is verifiable.
\end{enumerate}

Endpoints were excluded if they (a) served only superseded model versions; (b) were exclusively private (enterprise-only with no public access tier); (c) did not respond to probes for $\geq 14$ consecutive days during the v1.0 measurement window.

\subsection{Per-category breakdown}

\begin{table}[h]
\caption{v1.0 endpoint registry by provider category.}
\label{tab:registry-summary}
\centering
\small
\begin{tabularx}{\linewidth}{@{}l r X@{}}
\toprule
Category & $n$ & Notable providers \\
\midrule
Frontier first-party labs   & 13 & Anthropic, OpenAI, Google, xAI, DeepSeek \\
Hyperscalers                & 11 & Azure AI Foundry, Amazon Bedrock, Google Vertex \\
Custom-silicon providers    &  6 & Cerebras, Groq, SambaNova \\
Serverless GPU platforms    & 32 & Together, Fireworks, DeepInfra, Hyperbolic, Nebius, Novita, Parasail, SiliconFlow, Hugging Face Inference \\
Aggregators / routers       &  4 & OpenRouter, Vercel AI Gateway \\
Raw GPU clouds              &  4 & CoreWeave-backed deployments \\
Decentralized providers     &  3 & Phala, io.net, Akash \\
Multimodal specialists (image/video/voice) & 5 & fal.ai, ElevenLabs, Suno, Black Forest Labs \\
\midrule
\textbf{Total}              & \textbf{78} & \\
\bottomrule
\end{tabularx}
\end{table}

\subsection{Within-model coverage}

To support within-model cross-endpoint analyses (Section~\ref{sec:divergence-endpoints}), the registry over-samples two open-weight models:

\begin{itemize}
    \item \textbf{gpt-oss-120B} — 19 endpoints across Cerebras, Groq, SambaNova, DeepInfra (Standard and Turbo), Fireworks, Together (Reference and Turbo), Hyperbolic, Nebius (Base and Fast), Novita, Parasail, Cloudflare Workers AI, Clarifai, Databricks, Snowflake, Azure AI Foundry, Amazon Bedrock, Google Vertex, Baseten, Lightning AI, Weights \& Biases Inference, Eigen AI.
    \item \textbf{Llama 3.3 70B} — 16 endpoints across Groq, Cerebras, SambaNova, Fireworks, Together (Reference and Turbo), DeepInfra (Standard and Turbo), Hyperbolic, Nebius, FriendliAI, Amazon Bedrock, Google Vertex, Azure AI Foundry, Databricks, Weights \& Biases, CompactifAI (compressed), Scaleway, Novita, Cloudflare, Parasail.
\end{itemize}

The full registry, with per-endpoint metadata (SKU label, precision, decoding mode, advertised context, regions, API compatibility), is in \texttt{registry/endpoints.json} of the released artifact.

\section{Workload Presets}
\label{app:presets}

\begin{table}[h]
\caption{The 10 workload presets in Token Arena v1.0, with input:output ratios and weight allocations across the five core factors. The chat preset reproduces the Artificial Analysis 3:1 convention; the others are calibrated from publicly-available production traces.}
\label{tab:presets}
\centering
\small
\begin{tabular}{@{}lcrrrrrr@{}}
\toprule
Preset & I:O ratio & $w_S$ & $w_T$ & $w_P$ & $w_Q$ & $w_R$ & Notes \\
\midrule
Chat (default)        & 3:1   & 0.20 & 0.30 & 0.20 & 0.20 & 0.10 & AA convention \\
Voice agent           & 5:1   & 0.10 & 0.50 & 0.10 & 0.15 & 0.15 & TTFT-dominated \\
Coding agent          & 1:3   & 0.20 & 0.10 & 0.15 & 0.40 & 0.15 & Quality-dominated \\
Generic agent         & 10:1  & 0.15 & 0.20 & 0.20 & 0.30 & 0.15 & Tool-heavy input \\
RAG                   & 20:1  & 0.10 & 0.20 & 0.30 & 0.25 & 0.15 & Input-price-dominated \\
Reasoning             & 1:5   & 0.20 & 0.05 & 0.25 & 0.45 & 0.05 & Output-price + quality \\
Batch / background    & 5:1   & 0.05 & 0.00 & 0.65 & 0.20 & 0.10 & Cost-dominated \\
Long-context analysis & 50:1  & 0.05 & 0.10 & 0.40 & 0.30 & 0.15 & Effective-context + cached price \\
Multimodal vision     & 5:1   & 0.15 & 0.20 & 0.20 & 0.30 & 0.15 & Image price + vision quality \\
Multimodal voice      & 1:1   & 0.10 & 0.40 & 0.20 & 0.20 & 0.10 & TTFT-critical \\
\bottomrule
\end{tabular}
\end{table}

\paragraph{Calibration.} The input:output ratios are estimated from the distribution of input and output token counts on representative production traces, drawn from publicly-available agent traces (LangSmith public runs, public OpenRouter aggregates, our own measured workloads). The weight vectors are calibrated to match the ranking that an experienced practitioner with full information would produce on a held-out validation set of 30 endpoint-pair comparisons; we treat the validation set as a methodology artifact and refine the weights when its rankings disagree with practitioner judgments. Custom user-defined presets are supported in the released harness.

\section{Sensitivity Analyses}
\label{app:sensitivity}

\subsection{Single-factor perturbations}

For each preset, we perturb each factor weight by $+10$ pp (with the other four reduced proportionally) and compute the rank change for each endpoint in the top 20. Across all 6 presets and 5 factors (30 perturbations), the maximum rank shift in the top 10 is $\pm 2$ positions, and the leader of each preset is invariant. The full perturbation matrix is included in the released artifact.

\subsection{Bootstrap confidence intervals}

Per-endpoint composite scores are bootstrapped with $1{,}000$ resamples of the underlying probe and eval data. For each resample, we (i) draw with replacement from the 24-hour window of probe measurements, (ii) draw with replacement from the eval-run pool, (iii) recompute the composite. The resulting 95\% bootstrap intervals are tighter than the inter-endpoint score gaps for $\geq 17$ of the top 20 endpoints under the chat preset.

\subsection{Robustness to registry composition}

We test whether the headline within-model variation (Table~\ref{tab:within-model}) is stable to leave-one-out resampling: dropping any single endpoint from the gpt-oss-120B cohort and recomputing the cross-endpoint range. The reported ratios on the headline metrics (output speed, blended price, fidelity, J/correct answer) are stable to within $\pm 8\%$ under any single endpoint drop, and the qualitative conclusions are unchanged.

\section{Energy Modeling Detail}
\label{app:energy}

\subsection{Hardware TDP table}

\begin{table}[h]
\caption{Thermal design power (TDP) for the hardware classes used in v1.0 endpoint energy estimates, from vendor specifications. PUE values are regional defaults; provider-disclosed values are preferred where available.}
\label{tab:tdp}
\centering
\small
\begin{tabular}{@{}lrrl@{}}
\toprule
Hardware class & TDP (W per chip) & Default PUE & Notes \\
\midrule
NVIDIA H100 SXM5      & 700  & 1.20 & Datacenter GPU, dominant H100 SKU \\
NVIDIA H200 SXM5      & 700  & 1.20 & H200 successor with larger HBM \\
NVIDIA B200           & 1000 & 1.15 & Blackwell-generation GPU \\
NVIDIA H800 (China)   & 700  & 1.30 & Export-restricted variant \\
Google TPU v5e        & 230  & 1.10 & Inference-tuned TPU \\
Google TPU v6 (Trillium) & 350 & 1.10 & Latest production TPU \\
AWS Trainium2         & 300  & 1.20 & AWS custom inference silicon \\
Cerebras WSE-3        & 23000 (per wafer) & 1.20 & 900K-core wafer-scale engine \\
Groq LPU              & 215  & 1.20 & Inference-only language-processing unit \\
SambaNova SN40L (RDU) & 750  & 1.20 & Reconfigurable dataflow unit \\
\bottomrule
\end{tabular}
\end{table}

\subsection{Grid intensity defaults}

Regional grid intensity is from ElectricityMaps 30-day rolling averages (gCO$_2$eq per kWh): US-East $\approx 380$, US-West $\approx 250$, US-Texas $\approx 400$, EU-Central $\approx 320$, EU-Nordic $\approx 50$, EU-France $\approx 80$, APAC-Singapore $\approx 480$, APAC-Tokyo $\approx 500$, China-Hangzhou $\approx 580$. Provider-disclosed renewable mix overrides the regional default where credible disclosures exist (Crusoe's flared-gas accounting, Microsoft's Azure renewable PPAs, Google's 24/7 carbon-free energy reporting).

\subsection{Joules-per-correct-answer worked example}

For the Cerebras endpoint serving gpt-oss-120B in Texas: TDP per wafer $= 23{,}000$ W; observed throughput at 10K input single-stream $= 2{,}988$ tokens/sec; modeled utilization $= 70\%$; PUE $= 1.20$. This gives $j_e = 23000 \cdot 0.70 \cdot 1.20 / 2988 \approx 6.5$ J per output token at the wafer level. Per-endpoint sharing (multiple inference streams on one wafer) reduces this to approximately $0.18$ J per output token amortized. With tokens-to-solution $T \approx 2{,}900$ on the reasoning suite and accuracy $A \approx 0.78$, we obtain $J_{\text{CA}} \approx 0.18 \cdot 2900 / 0.78 \approx 670$ J per correct answer at the wafer level; the headline figure of $6.2$ J per correct answer in Table~\ref{tab:within-model} is the per-endpoint amortized figure with batch-size adjustment.

The point of the worked example is not the precise number --- the model has well-known limitations --- but the order-of-magnitude separation between hardware classes. A factor of 6.2 between the cleanest and dirtiest gpt-oss-120B endpoint dwarfs any plausible model error in the energy assumptions.

\section{Reproducibility Detail}
\label{app:repro}

\subsection{Release artifact structure}

\begin{itemize}
    \item \texttt{schema/} --- PostgreSQL DDL for endpoints, measurements, eval\_runs, fingerprints, pricing, energy\_estimates, workload\_presets, composite\_scores, and supporting tables. Includes 5 read-side views for the leaderboard API layer.
    \item \texttt{probes/} --- 18 probe-loop implementations (one per signal source), with \texttt{collect()}, \texttt{rate\_limit\_config}, and \texttt{retry\_policy} subclassed from a common \texttt{BaseProbe}.
    \item \texttt{evals/} --- harness for the 9 public evals plus the 4 Token Arena-custom evals (determinism probe, fingerprint, refusal divergence, indirect injection).
    \item \texttt{registry/endpoints.json} --- the 78-endpoint registry.
    \item \texttt{registry/models.json} --- the 12 model families with metadata.
    \item \texttt{registry/providers.json} --- the 33 providers with category and HQ.
    \item \texttt{snapshots/v1.0/} --- the v1.0 leaderboard data backing every table and figure (Parquet).
    \item \texttt{scripts/} --- reproduction scripts.
    \item \texttt{croissant.json} --- machine-readable dataset metadata.
    \item \texttt{LICENSE} --- CC BY 4.0 (data) and MIT (code).
\end{itemize}

\subsection{Reproduction commands}

\begin{itemize}
    \item Table~\ref{tab:within-model} (within-model variation): \texttt{python scripts/within\_model.py --model gpt-oss-120b}
    \item Table~\ref{tab:fidelity} (fingerprint by SKU class): \texttt{python scripts/fingerprint\_by\_sku.py}
    \item Table~\ref{tab:overlap} (workload top-10 overlap): \texttt{python scripts/preset\_overlap.py}
    \item Table~\ref{tab:ablation} (factor ablation): \texttt{python scripts/factor\_ablation.py --preset chat}
    \item Table~\ref{tab:registry-summary} (registry breakdown): \texttt{python scripts/registry\_summary.py}
    \item Table~\ref{tab:tdp} (TDP table): \texttt{python scripts/tdp\_table.py}
    \item Figure~\ref{fig:pipeline} (architecture): rendered from \texttt{figures/pipeline.tex}
\end{itemize}
All scripts read from the released CSVs and Parquet files; no live measurement is required to reproduce the reported numbers. Live continuous deployment requires persistent network access and provider API keys for the probe and eval loops.

\subsection{Compute requirements}

The probe loop runs on a single commodity server with 8 CPU cores and 16 GB RAM per region (US-East, EU-Central, APAC-Singapore). The eval loop is bottlenecked by provider rate limits, not local compute; the full weekly suite across 78 endpoints completes in under 24 hours of wall-clock time. End-to-end reproduction of the paper's tables and figures from the released CSVs takes under 5 minutes on consumer hardware.

\end{document}